\def\year{2022}\relax
\documentclass[letterpaper]{article} % DO NOT CHANGE THIS
\usepackage{aaai22}  % DO NOT CHANGE THIS
\usepackage{times}  % DO NOT CHANGE THIS
\usepackage{helvet} % DO NOT CHANGE THIS
\usepackage{courier}  % DO NOT CHANGE THIS
\usepackage[hyphens]{url}  % DO NOT CHANGE THIS
\usepackage{graphicx} % DO NOT CHANGE THIS
\usepackage{natbib}  % DO NOT CHANGE THIS AND DO NOT ADD ANY OPTIONS TO IT
\usepackage{caption} % DO NOT CHANGE THIS AND DO NOT ADD ANY OPTIONS TO IT
\usepackage{array, makecell}

\usepackage{booktabs}
\usepackage{multirow}
\usepackage{amsmath}
\usepackage{amssymb}
\usepackage{subfig}
\usepackage{todonotes}
\usepackage{tabularx}
\title{Beyond Toxic: Toxicity Detection Datasets are Not Enough for Brand Safety}
\author {
    Elizaveta Korotkova\textsuperscript{\rm 1}\footnote{Work done during internship at Clarifai},
    Isaac Kwan Yin Chung\textsuperscript{\rm 2}\\
}
\affiliations {
    \textsuperscript{\rm 1} University of Tartu  
    \textsuperscript{\rm 2} Clarifai\\
    lisa\_k@tartunlp.ai, first.last@clarifai.com
}
\begin{document}

\maketitle

%TODO-new: !!!!
\begin{abstract}
The rapid growth in user generated content on social media has resulted in a significant rise in demand for automated content moderation. Various methods and frameworks have been proposed for the tasks of hate speech detection and toxic comment classification. In this work, we combine common datasets % across tasks and use weighted sampling 
to extend these tasks to brand safety. %, i.e. identifying contexts where advertisements should not appear to protect commercial branding. This task covers not only toxicity, but also other potentially harmful content.
Brand safety aims to protect commercial branding by identifying contexts where advertisements should not appear and covers not only toxicity, but also other potentially harmful content.
%explore methods to extend these tasks to brand safety by combining datasets across tasks and weighted sampling.
As these datasets contain different label sets, we approach the overall problem as a binary classification task.
We demonstrate the need for building brand safety specific datasets via the application of common toxicity detection datasets to a subset of brand safety and empirically analyze the effects of weighted sampling strategies in text classification.
% We demonstrate that it is possible to achieve improved performance by the proposed methods.
% Empirical results confirm the gap between the brand safety domain and that of existing public toxicity detection datasets to be significant. The zero-shot abilities of public toxicity detection models on brand safety data reveal the need for building specific brand safety datasets.
\end{abstract}

\section{Introduction}
%\todo[inline]{a lot of UGC; need to automate content moderation; data for a particular domain is scarce and generalization is hard; but we do have some published datasets; what are those? can we use them? let's find out!}

As the amount of user generated content on social media continues to grow, a need to automate content moderation to address the scale problem has arisen, and many turn to using Deep Learning models and frameworks. In particular, harmful sources like hate speech and toxic comments have been targeted for moderation \cite{measurement_hs, cyberbullying-in-games}. 
Thus, various and often overlapping text moderation tasks, such as Hate Speech Detection, Toxic Comment Classification, and Offensive Language Detection \cite{basile-etal-2019-semeval, Fortuna-2018,ranasinghe-zampieri-2020-multilingual}, have received increasing attention from the NLP research community. Here we will refer to the broad task as toxicity detection.
 % Moreover, it was proposed to use a hierarchical approach to analyze different aspects, such as the type and the target of the offense, which helps provide explainability. However this is still an area under constant development and is susceptible to many specificities. % some bridging sentence to talk about why we don't use the hierarchical labels.
% A similar hierarchy was proposed \cite{Alkomah_lit_review} where hate speech is seen as a layer between aggressive and abusive text; however, all of these share the offensive aspect. 

Deep learning models often perform well on specialized datasets %for specific tasks 
when they are evaluated on their respective train-test splits;  however, these models are unable to generalize to new hate speech content \cite{Wullach, Arango2019-sigir}. Datasets often differ in their class definitions and labelling criteria, leading to poor model performance when predicting across new domains and datasets. One way to address this problem is to collect more data from the target domain, but in reality quality data is often scarce. Thus, a common technique is to include additional data from related sources. 

Brand safety is an industry application of content moderation that has received increasing attention. It refers to understanding and identifying contexts in which advertisements should not appear.\footnote{In this work, we consider the textual aspect of brand safety, while acknowledging that brand safety is typically evaluated in a multi-modal setting.} 
While hate speech and toxic comments are certainly part of most brand safety applications, other moderation criteria such as spam, crime, and illegal drug use are also present in the widely adopted GARM Brand Safety Floor and Suitability Framework \cite{wfa_2022}.
Although the need to automate brand safety is undeniable, this field remains under-explored in literature as there are, to our knowledge, no existing publications or current benchmarks.

We perform an initial case study of methods that could leverage existing datasets to generalize well in a different and potentially wider domain such as brand safety. We apply these methods in a binary classification problem to classify text into offensive and non-offensive categories. 
Our contributions are:
\begin{itemize}
    \item We demonstrate the need for building brand safety specific datasets via the application of common toxicity detection datasets to a subset of brand safety.
    \item We empirically analyze the effects of weighted sampling strategies in text classification.% by evaluating on publicly available test sets and a private brand safety dataset.
    % \item We expose challenges associated with label definitions of public text moderation datasets in a binary text classification setting, specifically in the context of brand safety.
\end{itemize}

\section{Related Work}
\subsection{Toxicity Detection}
Works on toxicity detection often cover a large variety of domains such as social media platform contents \cite{solid}, news comments \cite{korencic-etal-2021-block} and hate speech from online forums \cite{de-gibert-etal-2018-hate}. There are also works that focus on culture-, region- or language-specific applications, such as \citet{singapore} and \citet{Li_Ning_2022}. %Then somewhat related data augmentation methods like unsupervised stuff. Then looking at robust route. 
 Arango et al. \citeyearpar{Arango2019-sigir}  showed  that  cross-dataset  performance  can  be  improved  by  removing  bias from the training data, and adding data from another source.
% (in  this  case  the  hate  speech  data  provided  by  Davidson, Bhattacharya, and Weber (2019)). However, it is not clear whether the performance gain achieved by Arango et al. is caused by de-biasing or augmenting the data. Moreover, this cross-dataset experimentation was not complete. Typically domain-adaptation studies evaluate a model trained from a source domain across more than one target domain. % copied from https://ojs.aaai.org/index.php/ICWSM/article/view/19340/19112 so needs rewording
Waseem et al. \citeyearpar{Waseem2018BridgingTG}  proposed  a  multi-task learning approach to integrate different datasets into a single training process to construct a generalized hate speech detection model. % same as above. needs rewording. Ref: Waseem, Z.; Thorne, J.; and Bingel, J. 2018.Bridging theGaps:  Multi  Task  Learning  for  Domain  Transfer  of  HateSpeech  Detection,  29–55.Cham:  Springer  InternationalPublishing. ISBN 978-3-319-78583-7.
% \item robust fine-tuning of zero-shot models: shows visualization methods to understand ppOOD/ID between datasets to quantify domain similarity %https://rosanneliu.com/dlctfs/dlct_211025.pdf %% we can reference this when presenting the dataset correlation matrix
% \citet{Sarwar_Murdock_2022} augment weakly labeled data in an unsupervised manner for domain adaptation. %This is likely the most similar to our work. 
Multi-task learning \cite{Waseem2018BridgingTG, Yuan} has been previously used to improve generalization of hate speech detection models. \citet{Yuan} also binarize the class labels from multiple datasets into harmful vs non-harmful labels.

\subsection{Weighted Sampling}
%\todo[inline]{weighted sampling based on labels for imbalanced classification. ect etc etc}
% we understand datasets; could it tell us how a different dataset is labeled? could we infer
% Some prior works use instance level weighting.
% https://arxiv.org/pdf/1812.03372.pdf
Previous works have shown empirically that various sampling techniques can improve classification of imbalanced image data \cite{Masko2015TheIO, Byrd2018WhatIT} and imbalanced text data \cite{Padurariu}. Oversampling has been shown to be effective in overcoming the class imbalance in SemEval top submissions \cite{offenseval-2020}. In this work, we explore weighted sampling by class and datasets (when multiple datasets are combined).% in this application.
% TODO-new sentence that relates these to your work --- e.g. about explore sampling strategies in this application (?) 

% Its effect in text moderation, where classes are inherently heavily imbalanced, remains under-explored. 
% \todo{I'm not sure we can say that; e.g. "they used a very simple approach: over-sample the training data to overcome the class imbalance" in https://aclanthology.org/2020.semeval-1.188.pdf} 
% maybe we can say we extend their approach, or simply remove this sentence?

\section{Data}
% TODO: Here we can use the space to mention some findings from the data exploration, such as labeling criteria differences: e.g. some consider profanity as offensive, some consider the context in which it was used (surge). Some had emojis?
% For the purposes of practical applicability and reproducibility, we only use publicly available datasets under permissive licences for training our models. 
In this section, we describe the datasets used in our experiments. Aiming to explore the effectiveness of existing toxicity detection datasets, we focus on readily available datasets distributed under permissive licenses. Some challenges of using such data are their differing label sets and varying, often unclear, labeling criteria. Train and test data statistics are shown in Table \ref{tab:datasets}. 

% Old figure commented out below, these are balanced
% \begin{figure}[h]
%     \centering
%     % \subfloat[\centering Pre-trained DistilBERT]{{\includegraphics[width=6cm]{sample_similarities_heatmap_balanced.png} }}
%     % \qquad
%     % \subfloat[\centering Finetuned Model trained on combined data]{{\includegraphics[width=6cm]{sample_similarities_heatmap_balanced_finetuned.png} }}
%     \includegraphics[width=8.5cm]{sample_similarities_heatmap_balanced_finetuned.png}
%     \caption{We use each model to embed 400 positive and 400 negative samples for each test set. Cosine similarities between the averaged embeddings are plotted as a heatmap.}%\todo[inline]{bigger font in plot?}
%     \label{fig:example}
% \end{figure}

%%% PLACEHOLDER PLOT: TODO FOR LISA
\begin{figure}[h]
    \centering
    % \subfloat[\centering Pre-trained DistilBERT]{{\includegraphics[width=6cm]{sample_similarities_heatmap_balanced.png} }}
    % \qquad
    % \subfloat[\centering Finetuned Model trained on combined data]{{\includegraphics[width=6cm]{sample_similarities_heatmap_balanced_finetuned.png} }}
    \includegraphics[width=8.5cm]{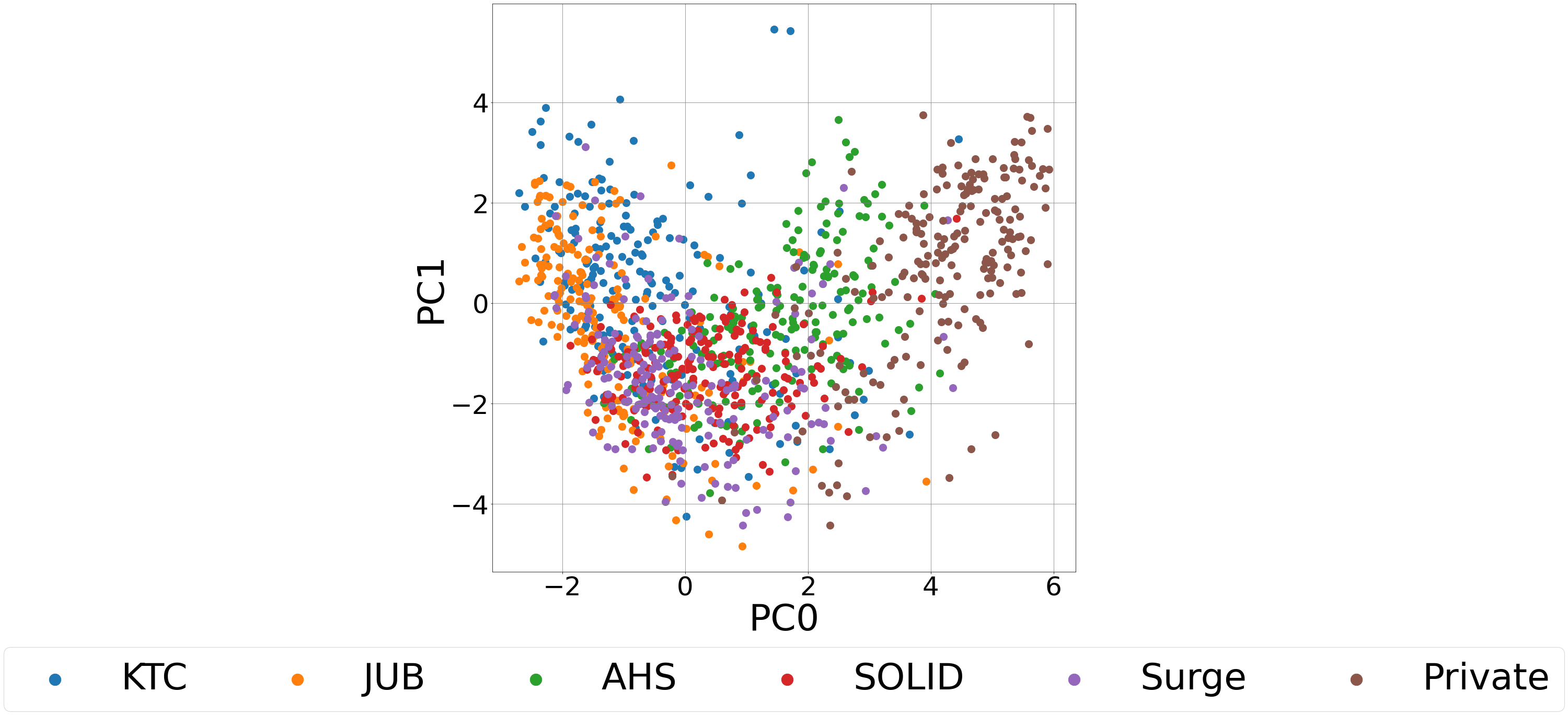}
    \caption{PCA projection of embeddings extracted from the 4th Transformer layer of the pre-trained DistilBERT model (200 examples sampled from each dataset). Each example's embedding is the average of its token embeddings. }%\todo[inline]{bigger font in plot?}
    \label{fig:scatterplot}
\end{figure}

\begin{centering}
\begin{table}
\setlength\tabcolsep{3pt}
\begin{tabularx}{\columnwidth}{lcccc}
% \begin{tabular}{lp{0.15\textwidth}cccc}
\toprule
\thead{Dataset} & \thead{\#Train} & \thead{\#Test} & \thead{*\#Positives} & \thead{*AvgTextLen} \\ 
  \midrule
AHS** & 22,305 & **2,478 & 2,062 & 85.4 \\
SOLID & - & 3,887 & 1,080 & 78.1 \\
Surge & - & 1,000 & 501 & 135.6 \\
KTC & 159,571 & 63,978 & 6,243 & 383 \\
JUB & 1,804,874 & 97,320 & 29,052 & 296 \\
Private & - & 37,352 & 6,534 & 81.3 \\
\bottomrule
% \end{tabular}
\end{tabularx}
\caption{\label{tab:datasets} Description of the toxicity detection and brand safety datasets. Average text length is measured in characters. \\ Note *: For test sets. **: train-test split was 90-10.}
\end{table}
\end{centering}

% \subsection{Public Datasets}
\subsection{Common Toxicity Detection Datasets}
% TODO-new  begin each paragraph with the name of the dataset
``Automated Hate Speech Detection and the Problem of Offensive Language'' by \citet{automated_hs} (AHS) is comprised of examples where 77.4\% are categorized as \textit{offensive language}, 5.8\% as \textit{hate speech}, and 16.8\% as \textit{neither}.
% The dataset, further referred to as AHS, contains 24,783 non-anonymized tweet texts, distributed under the MIT license.
Annotators were instructed that the presence of a particular word, however offensive, did not necessarily indicate that a tweet is hate speech. Labels are assigned according to the majority decision.
%\todo[inline]{mention that no clear labeling definitions?}

Semi-Supervised Offensive Language Identification Dataset (SOLID) \cite{solid} was created for the SemEval-2020 task on Multilingual Offensive Language Identification in Social Media \cite{offenseval-2020}. 
% The data is collected from Twitter and is distributed under CC BY 4.0 license.
% The taxonomy of the dataset is hierarchical with 3 levels: at level A, the text is categorized as either offensive or not offensive; at level B, the offensive instances are marked as either targeted or untargeted; at level C, the target of the offensive language is identified (individual, group, or other). 
The labelling is hierarchical, but we only operate at the highest level and consider its two classes: \textit{NOT} (neither offensive, nor profane), and \textit{OFF} (inappropriate language, insults, or threats).
% The level A training set contains 9,089,140 instances, and the test set 3,887 instances. For the training set, only tweet IDs are provided, which means that using it would require retrieving the texts using an API, and the eventual number of usable instances will be much lower (because of deleted tweets, accounts that became private since the data was collected, etc.). In the test set, tweet texts are provided, with mentioned usernames anonymized as \texttt{@USER}.
% The training set was labeled in a semi-supervised way. Several classifiers were trained on OLID \cite{olid}, a smaller, supervised dataset, and the average prediction of those models between 0.0 and 1.0 is given as the label (standard deviation of the predictions is also provided). 
%We only use the test set, where the semi-supervised labels were validated by human annotators (one annotator per example). 
The dataset is imbalanced, with %around 
84\% of %level A 
training labels below 0.5.

Surge AI free Toxicity dataset\footnote{https://github.com/surge-ai/toxicity} is a sample of a larger, commercially distributed set\footnote{https://www.surgehq.ai/datasets/toxicity-dataset}. The data is from ``a variety of social media platforms''. 
Annotators had ``to identify comments that they personally found toxic'' without a strict definition of toxicity. Due to this, use of profanity by itself does not mean that an example is toxic, and intended meaning is taken into account.
% The sample contains 1000 lines (501 \textit{Toxic} and 499 \textit{Not Toxic}).

Toxic Comment Classification Challenge dataset\footnote{https://www.kaggle.com/competitions/jigsaw-toxic-comment-classification-challenge/data} (KTC) was published within a Kaggle competition. % and is distributed under the CC0 license. 
The data is sourced from Wikipedia comments and labeled by human raters for toxic behavior.
% There are 6 toxicity classes in the dataset: \textit{toxic}, \textit{severe\_toxic}, \textit{obscene}, \textit{threat}, \textit{insult}, \textit{identity\_hate}. %The class labels are binary (0 or 1). 
The dataset is multi-labeled using 6 toxicity classes, with an additional negative class that can be defined for non-toxic texts. %, where all positive class labels are 0. 
The training set is imbalanced: 89.8\% of instances are negative. 
%We use the same test set which was used for scoring the competition.

Jigsaw Unintended Bias in Toxicity Classification\footnote{https://www.kaggle.com/competitions/jigsaw-unintended-bias-in-toxicity-classification/data} (JUB) was also published within a Kaggle competition and was sourced from the Civil Comments platform. 
%Annotators rated the toxicity of each comment. 
The target toxicity label is between 0.0 and 1.0, showing what fraction of annotators marked the instance as either toxic or very toxic.
%Only the toxicity label was the prediction target within the competition, but 
The dataset also contains multi-class annotation similar to that of KTC. For each of the toxicity subtypes, a label between 0.0 and 1.0 is provided. %We consider 0.5 as the threshold to determine class membership.
The training set is imbalanced: 92\% of the data has a toxicity label below 0.5.

\subsection{Private Dataset}
\begin{figure}[h]
    \centering
    \includegraphics[width=8cm]{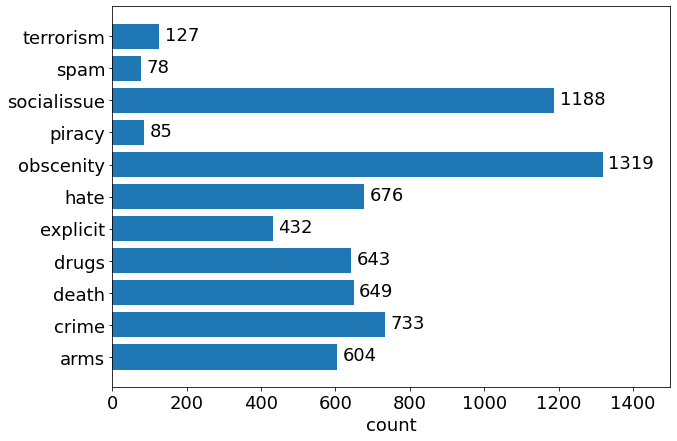}
    \caption{Counts of brand safety classes in the Private dataset. The dataset is multi-labeled, i.e. one example may belong to more than one class. Additionally, there are 29,596 negative examples in the dataset (not shown here).}
    \label{fig:classes}
\end{figure}

This dataset was sampled from a real-world customer %text moderation 
application in brand safety text moderation.
It consists of 55,755 user-generated text captions posted alongside videos on a social media platform. We only use the 37,352 English examples in our experiments. Annotators were instructed to indicate all brand safety classes that are present. Positive class counts are shown in Figure \ref{fig:classes}. %The obscenity class has the largest proportion at 1319 examples, followed by social issues at 1188, crime 733, hate speech 676, death 649, drugs 643, arms and weapons 604, explicit 432, terrorism 127, piracy 85, and spam 78. 
See Table \ref{tab:private-examples} for examples.

\begin{centering}
\begin{table}
\setlength\tabcolsep{4pt}
\begin{tabularx}{\columnwidth}{lp{7.5cm}}
\toprule
\multirow{2}{*}{Pos} & Yikes....(this was pre covid btw) \#racist \#homophopic \#exposed \#fyp \#foryou \#awkward \#lgbtq \#blm\\ \cmidrule{2-2}
 & oomf.. u caused this \#danganronpa \#byakuyatogami \#makotonaegi \#yasuhirohagakure \#genocidersyo \#aoiasahina \\
\midrule
\multirow{2}{*}{Neg} & im carrying this family on my back???? \\ \cmidrule{2-2}
%Neg & This is so satisfying go check out my slime shop link in bio ���� do y’all want to see more of this￼￼￼ \#slime \#satisfying \#foryou \#follow\\
 & \#greenscreensticker \#swordsman000 \#animethings \#hiddenanimesociety\\
\bottomrule
\end{tabularx}
\caption{\label{tab:private-examples} Examples from the Private dataset}
\end{table}
\end{centering}

\section{Methodology}

\subsection{Datasets}
We use the AHS, KTC, and JUB datasets to train our models. We implement two settings: 1) train a model using a single dataset, and 2) concatenate all three datasets to form the combined training set. We randomly select 10\% of each dataset for validation. 
We evaluate our models on all test sets. The full Surge dataset sample is used for testing due to its small size. 
For SOLID, we do not collect the training examples (for which only tweet IDs are provided) and use the provided test split %, where the semi-supervised labels were validated by human annotators, 
for evaluation. 
AHS does not have a designated test split, and we report results on its validation split. For KTC and JUB, we use the pre-defined training and test splits. The private dataset is treated as unseen and only used for testing. We extract an additional subset from it by combining negative (safe) examples and examples labelled as containing hate speech or obscenity, while excluding other brand safety classes.

\subsection{Preprocessing}

As the datasets have different classes and labeling criteria, we unify them by treating the task as a binary classification problem. We map all labels corresponding to hateful, toxic, and otherwise offensive texts to offensive (positive) class, and the rest to non-offensive (negative) class. If a value between 0.0 and 1.0 is provided instead of a discrete class label, we consider instances with toxicity scores strictly below 0.5 negative, and the rest positive. We do not apply any additional cleaning or filtering to the datasets.

\subsection{Models}

To obtain our classification models, we fine-tune the pre-trained DistilBERT model \cite{Sanh2019} using the Hugging Face Transformers library \cite{HFTransformers}. %Focusing on data rather than training techniques, 
We fine-tune each model for 3 epochs, keeping all other hyperparameters at their default values. For each model, we do 3 restarts of fine-tuning and report the average scores. We use 3 NVIDIA GeForce GTX TITAN X GPUs for each training run with a batch size of 16 instances per GPU.

% \subsection{Binarizing class labels and Resampling}
\subsection{Weighted Sampling}
%Since the chosen datasets have different taxonomies and labeling criteria, we binarize the class labels to map all hateful or offensive-related labels to be offensive (positive class), otherwise non-offensive. 
A strong class imbalance is seen across all training datasets, and the datasets differ significantly in size. This suggests that weighted sampling of instances during training may significantly influence the results shown by the models. In the default setting, we combine all training datasets into one dataset and sample its instances with equal probability. Additionally, we implement 3 variants of weighted sampling: with an equal probability of sampling from each of the 2 classes (to account for class imbalance), from each of the 3 datasets (to account for different datasets being unequally represented), and with equal weight given to each class of each dataset (the two previous strategies combined).

% We show that the proposed methods above improves results across all test sets by comparing results from models trained on each individual dataset (with exceptions to SOLID and Surge). 
% For all models, we fine-tune DistilBERT for 3 epochs for time considerations.
% We use 2 NVIDIA GeForce GTX TITAN X GPUs for training each model, with a batch size of 16 instances per GPU. 
% We keep all other hyper-parameters as they are in the default. We report precision, recall and F1 score as performance metrics and their respective values averaged over 3 runs.

\subsection{Evaluation}

For each of the models, we report %precision, recall, and 
the F1-score on each of the test sets, averaged over 3 training runs. 

%% F1 score only table with individual datasets
\begin{centering}
\begin{table}[h]
\centering
\setlength\tabcolsep{4pt}
\begin{tabular}{ccccccccc}
\toprule
\multirow{2}{*}{Test set} & \multicolumn{3}{c}{Training set}\\
\cmidrule{2-4}
 & KTC & JUB & AHS \\ \midrule
KTC &\textbf{ 0.674} & 0.666 & 0.592\\
JUB & 0.508 & \textbf{0.691} & 0.296 \\
AHS & 0.897 & 0.846 & \textbf{0.995} \\
SOLID  & 0.873 & \textbf{0.874} & 0.782 \\
Surge & \textbf{0.762} & 0.739 & 0.613  \\
Private &\textbf{0.253} & 0.135 & 0.189 \\
Private-Subset & \textbf{0.280} & 0.199 & 0.275  \\
\bottomrule
\end{tabular}
\caption{\label{tab:results-cross-dataset}  Cross-dataset performance in F1 scores. In most cases, the results are clearly worse on out-of-domain data. Highest results for each test set are shown in \textbf{bold}. For AHS, we report results on the validation set.}
\end{table}
\end{centering}

%% F1 score only table with only combined dataset + weighted sampling
\begin{centering}
\begin{table}[h]
\centering
\setlength\tabcolsep{4pt}
\begin{tabular}{cccccc}
\toprule
\multirow{2}{*}{Test set} & \multicolumn{4}{c}{Sampling strategy}\\
\cmidrule{2-5}
 & Combined & By label & By dataset & Both \\ \midrule
KTC &  \textbf{0.690} & 0.652 & 0.686 & 0.669\\
JUB &  \textbf{0.687} & 0.676 & 0.675 & 0.646 \\
AHS  & 0.989 & 0.989 & \textbf{0.998} & \textbf{0.998}  \\
SOLID  & \textbf{0.879} & \textbf{0.879} & 0.858 & 0.873 \\
Surge & 0.751 & 0.800 & 0.740 & \textbf{0.837}  \\
Private & 0.162 & 0.198 & 0.194 & \textbf{0.203}\\
Private-Subset & \textbf{0.273} & 0.240 & 0.261 & 0.223 \\
\bottomrule
\end{tabular}
\caption{\label{tab:results-dataset-weighted} F1 scores on different test sets achieved by fine-tuning on the combined training set with different weighted sampling strategies. "Combined" has no weighted sampling. "Both" samples by both label and dataset. Highest results for each test set are shown in \textbf{bold}. For AHS, we report results on the validation set.}
\end{table}
\end{centering}

\begin{figure}[h]
    \centering
    \includegraphics[width=7cm]{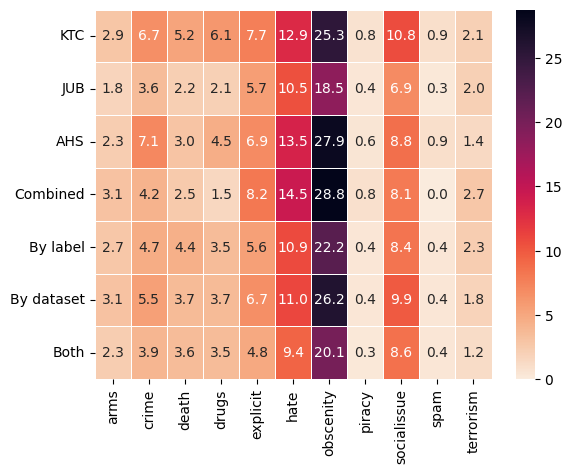}
    \caption{Heatmap of F1 scores $\times 100$ per class breakdown over all models on the Brand Safety test set. Obscenity and Hate Speech correspond to the toxicity detection domain better than the rest of the classes.}
    \label{fig:per-class-heatmap}
\end{figure}

\section{Results}
%% Individual sets
Results of models trained on different training sets are shown in Table \ref{tab:results-cross-dataset}. As expected, models perform best on their own respective test sets. %The cross-dataset drop in metrics confirms the presence of domain differences between the datasets. 
The model trained on KTC yields better results than other training sets on all test-only sets except for SOLID. %The KTC model is the best for the Surge test set according to all metrics. It also shows the best recall for all test-only sets and the best F1 score for Surge and Private. %, while its F1 for SOLID is a very close second. 
%% maybe consider removing the paragraph below to make room for discussion for the private set?
%At the same time, the AHS model yields higher recall for 2 of the 3 additional test sets. The JUB model seems to reach a reasonable balance between precision and recall to achieve the highest F1 score. 
%The best scores for the 3 metrics of the SOLID test set are achieved by different models.
However, on the Private test set, all models show a large drop in metrics.%, which suggests a significant extent of domain shift between the Private dataset and all others.

%Interestingly, we observe that using KTC yields better results than other training sets over the 3 additional test sets. This was unexpected since the test sets come from different data sources (see Table \ref{tab:datasets}). 
%shows results of our experiments where the models are trained either on a single dataset, or on all datasets combined. We can observe that performance on each of the training datasets benefits. shows the drop in recall, when the training and test set are from different datasets compared to when they are from the same dataset. Additional evaluations done on SOLID and Surge reveal that the model trained on KTC has the best zero-shot performance on this binary task.  \todo[inline]{describe table numbers in text here} 
Table \ref{tab:results-dataset-weighted} shows the performance of various weighted sampling strategies over the combined dataset on the test sets. 
Simply combining all datasets without balancing yields the highest F1 scores in 4 test sets. Balancing by both label and dataset yields highest metrics in 3 test sets. 
%No single strategy seems to yield dominating results. However, simply combining all training data allows us to outperform the best F1 score achieved by any single-dataset model for the KTC and SOLID test sets.
%The highest recall for 3 and the highest F1 score for 2 test-only sets in Table \ref{tab:results-dataset-weighted} are shown by the model trained with weighted sampling by label and dataset. 
%In general, weighted sampling of classes ("By label" and "Both") boosts recall, which is likely explained by the model encountering more positive instances during training. Simply combining all datasets without balancing leads to the best precision scores for 2 test-only sets. %Weighted sampling by dataset, as expected, improves the model's performance on the smallest training dataset, AHS.

\section{Discussion}
%the prev section seems all over the place for now, I want to separate everything and maybe combine later
In general, all fine-tuned models perform reasonably well on toxicity detection test sets. Weighted sampling is able to offer marginal boosts in performance.
% Training on JUB not only yields the highest metric on the JUB test set, but also on SOLID. This could be due to the test sets' high embedding similarity shown in Figure \ref{fig:heatmap}.
% \begin{centering}
% \begin{table}
% \setlength\tabcolsep{4pt}
% \begin{tabularx}{\columnwidth}{X|X}
% \toprule
% FP & FN \\
% \midrule
% For real ep 6 is what took the show from "pretty good" to "i fucking love this show" for me & Can’t wait to see you dressed in an orange jumpsuit\\
% \bottomrule
% \end{tabularx}
% \caption{\label{tab:fpfn-surge} FP and FN Examples from the Surge Dataset}
% \end{table}
% \end{centering}
\begin{centering}
\begin{table}
\setlength\tabcolsep{4pt}
\begin{tabularx}{\columnwidth}{lp{7.5cm}}
\toprule
FP & For real ep 6 is what took the show from "pretty good" to "i fucking love this show" for me \\
\midrule
FN & Can’t wait to see you dressed in an orange jumpsuit\\
\bottomrule
\end{tabularx}
\caption{\label{tab:fpfn-surge} False positive (FP) and false negative (FN) examples from the Surge dataset}
\end{table}
\end{centering}
Table \ref{tab:fpfn-surge} shows two challenging examples from the Surge dataset misclassified by the model fine-tuned on all data. %without weighted sampling. %We can immediately observe why these examples are hard to classify correctly. 
In the first example, profanity is used to convey a positive sentiment. While in the Surge dataset profanity does not mean the example will be labeled as toxic, in some other datasets it does. Such differences in labelling are undeniably problematic for our models. For the false negative, world knowledge (which is also culture-specific in this case) is needed to recognize the text as toxic/offensive. 
% Looking deeper into the results from the combined model reveal that often times presence of profanity leads to False Positives (FP). This is obvious in Surge where use of profanity does not mean that an example would be marked as toxic. On the other hand, some harmful/toxic comments that require a level of world knowledge often lead to False Negatives (FN). Examples of such FP and FN are shown in \ref{tab:fpfn-surge}. 
% - Inconclusive with SOLID performances. Could be due to the difference in machine/labeler consensus when labeling the SOLID test set.
% - Claims to "difficult" or "out of domain" examples? taking a lot of context into account sometimes. World-awareness needed.  (can refer back to some FP/FN)

%%% Similarity from heatmap
% From Figure \ref{fig:scatterplot} we observe that the embeddings of the Private dataset sample seem to form their own cluster, which suggests that Private is dissimilar from the public text moderation datasets.

%%% Applicability to private
% TODO-new: (this paragraph)
%   try some other things for this paragraph or rewrite and edit 

% could rephrase in a more positive light: link to original hypothesis, start with things that confirm it, then go to things that are different. 
Our hypothesis is that the hate speech and obscenity portions of the Private dataset would correspond the closest to the toxicity detection datasets, which is confirmed by higher F1 scores across all models on the Private-Subset than on the Private test set in Tables \ref{tab:results-cross-dataset} and \ref{tab:results-dataset-weighted}. From the per-class breakdown shown in Figure \ref{fig:per-class-heatmap}, we also observe that the aforementioned portions are indeed relatively closer compared to the rest of the GARM classes.
Debated social issues and explicit sexual content portions correspond the next closest to the toxicity detection datasets. 
These scores are still much lower compared to the results achieved on the other test sets. This suggests many further differences, and we observe this dissimilarity in Figure \ref{fig:scatterplot} as well.
In general, brand safety in text moderation remains an under-explored area without many prior publications, which prompted us to explore the applicability of the related task of toxicity detection. 
% TODO-new:  State findings here:  how did it apply? (yes but only for a portion)
Empirical results suggest that even though the two domains are related, this only applies to a small portion of brand safety aspects.
Additional data collection and continuing to align brand safety-focused data to the current toxicity definitions will benefit future work. 
This is important because harmful content moderation is frequently driven in part by the pragmatic desire to protect commercial branding, thus creating a need to understand both tasks and how they relate.
%Building a brand safety dataset would require data in domains other than offensive language detection. %Other considerations include the trade-off between precision and recall, which might be different for e.g. toxicity moderation and brand safety. 

\section{Conclusion and Future Work}

In this work, we perform an initial case study of extending toxicity detection to the task of brand safety. Our results reveal various challenges in using common toxicity detection datasets for solving this task, including domain differences between the datasets and a coverage mismatch between the tasks' label sets. We begin to address this gap using multi-dataset training and weighted sampling.

The scope of this work is limited to English data. %, and this limitation is very significant, especially when dealing with user-generated content. 
In future work, it could be extended to other languages by using multilingual large language models and exploring relevant available data. 
Methods for bridging the gap between different domains and tasks could also be explored and may include combining the toxicity detection and brand safety data in various ratios.

%More extensive data assembly for brand safety is also encouraged. 
\section{Social Impact}
% Research on brand safety can help broaden the moderation on social media in ways outside of hate speech detection and limit the impacts of harmful online content.
Research on brand safety can help in the moderation of potentially harmful content in addition to hate speech, including illegal drug use and illegal arms use, taking a step towards a safer digital media environment.
% [TODO exampels of 1 or 2 categories]
% TODO-new:  

\section{ Acknowledgments}
We thank David Eigen, Michael Gormish, and Eran Nussinovitch for their support and guidance in this study, and Karen Herder and Michael Tolbert for their support in gathering the private dataset.

\bibliography{bibliography.bib}

\end{document}